\documentclass[journal, twocolumn]{IEEEtran}
\IEEEoverridecommandlockouts
\usepackage{graphicx}
\usepackage{graphics}
\usepackage{xcolor}
\usepackage[skip=0pt]{subcaption}
\usepackage{amsmath}
\usepackage{gensymb}
\usepackage{amssymb}
\usepackage{algorithm}
\usepackage{algorithmicx}
\usepackage[noend]{algpseudocode}
\usepackage{booktabs}
\usepackage{titlesec}
\usepackage{listings}
\usepackage{enumitem}
\usepackage{multirow}
\usepackage{placeins}
\usepackage{rotating}
\usepackage{pifont}
\usepackage{array}
\usepackage{mdframed}
\usepackage{lipsum}
\usepackage[utf8]{inputenc}
\usepackage[colorlinks=true, allcolors=blue]{hyperref}
\usepackage[many]{tcolorbox}
\usepackage{varwidth}
\usepackage{cleveref}
\usepackage{colortbl}
\usepackage{array}
\usepackage{blindtext}
\usepackage{makecell}
\usepackage{subcaption}
\usepackage[font=scriptsize]{caption}
\usepackage{arydshln}
\usepackage{hhline}
\newcolumntype{?}{!{\vrule width 1pt}}
\newcolumntype{P}[1]{>{\centering\arraybackslash}p{#1}}

\usepackage{scalefnt}

\title{\LARGE RL-PGO: Reinforcement Learning-based Planar Pose-Graph Optimization \vspace{-8 mm}}
\author{
Nikolaos Kourtzanidis and Sajad Saeedi 
\vspace{-10 mm}
\thanks{Department of Mechanical and Industrial Engineering, 
Ryerson University, Toronto, Canada.
        {\tt\small \{nkourtza, s.saeedi\}@ryerson.ca}}%
}

\begin{document}
\maketitle

\begin{abstract}
The objective of pose SLAM or pose-graph optimization (PGO) is to estimate the trajectory of a robot given odometric and loop closing constraints. State-of-the-art iterative approaches typically involve the linearization of a non-convex objective function and then repeatedly solve a set of normal equations. Furthermore, these methods may converge to a local minima yielding sub-optimal results. In this work, we present to the best of our knowledge the first Deep Reinforcement Learning (DRL) based environment and proposed agent for 2D pose-graph optimization. We demonstrate that the pose-graph optimization problem can be modeled as a partially observable Markov Decision Process and evaluate performance on real-world and synthetic datasets. The proposed agent outperforms state-of-the-art solver $\mathrm{g}^{2} \mathrm{o}$ on challenging instances where traditional nonlinear least-squares techniques may fail or converge to unsatisfactory solutions. Experimental results indicate that iterative-based solvers bootstrapped with the proposed approach allow for significantly higher quality estimations. We believe that reinforcement learning-based PGO is a promising avenue to further accelerate research towards globally optimal algorithms. Thus, our work paves the way to new optimization strategies in the 2D pose SLAM domain.
\end{abstract}

\section{INTRODUCTION}
\label{intro}

 Bundle adjustment (BA) is a process that plays a significant role in computer vision applications today, such as Structure from Motion (SfM) and on the back-end of Simultaneous Localization and Mapping (SLAM) algorithms~\cite{BA-Revisited}. This process involves estimating the 3D coordinates describing the scene and pose of the camera concurrently. This estimation is typically performed by optimally minimizing a cost function which minimizes re-projection error. In large complex scenes, however, the number of landmarks are significantly more prominent, which will, in turn increase the computation time and may affect real-time performance. When converting the problem of this type, one can either avoid taking into account all historical data from previous time steps and apply a sliding window technique or eliminate the optimization of the landmarks~\cite{SLAMBOOK}. The latter is commonly referred to as pose-graph optimization. These relative pose measurements are typically obtained from cameras, laser sensors, IMUs, or wheel odometry. Given odometric or loop closing measurements at successive time intervals, the objective of PGO is to return the optimal configuration of pose estimates, which maximally explain the available measurements observed. The computation of the maximum likelihood estimate of robot poses results in a high-dimensional non-convex optimization problem with multiple local minima. 
 
State-of-the-art methods in today's modern era view SLAM and BA as Maximum a posteriori (MAP) or Maximum Likelihood inference. A specific class of graphs often analyzes both problems, known as factor graphs~\cite{FactorGraphBook}.

\par

\begin{figure}[t!]
 \centering
\subfloat[Initial noisy estimate of M3500B.]{\includegraphics[width=0.24\textwidth]{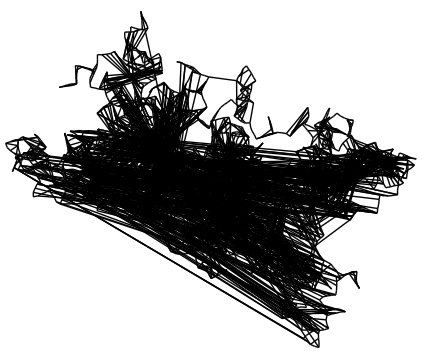}} 
\hfill
\subfloat[Proposed reinforcement learning agent's estimate]{\includegraphics[width=0.24\textwidth]{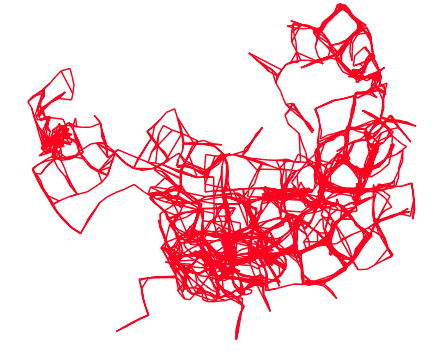}}
\hfill
\subfloat[$\mathrm{g}^{2} \mathrm{o}$'s Levenberg-Marquardt solver estimate after 100 iterations ]{\includegraphics[width=0.24\textwidth]{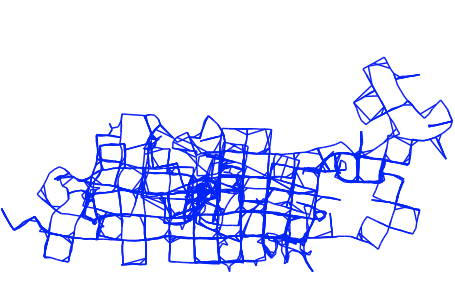}}
\hfill
\subfloat[Levenberg-Marquardt set to perform 30 iterations when bootstrapped by the proposed agent's result as an initial guess.]{\includegraphics[width=0.24\textwidth]{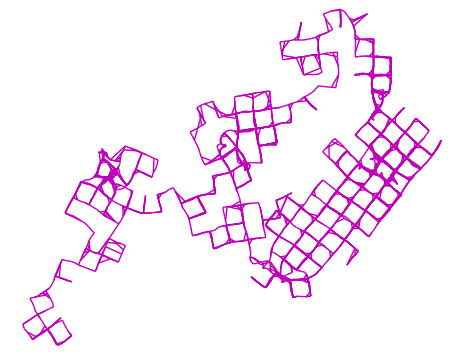}} 
\caption{Evaluation on the standard Manhattan world M3500B~\cite{carlone2012angular}. In this dataset, we notice that the estimate produced by Levenberg-Marquardt set to perform $100$ iterations yields an unsatisfactory solution. When bootstrapped by our proposed agent's initial guess, Levenberg-Marquardt set to complete just 30 iterations results in significantly improved accuracy.}
\vspace{-5 mm}
\label{fig:M3500B}
\end{figure}

There exist instances of SLAM problems which consist of highly non-convex cost functions, attributed to system nonlinearity, high levels of noise corruption, large inter-nodal distance spacing, and incorrect data associations from the front end~\cite{LAGO}. This further results in cost functions with {\it large valleys}~\cite{Olson2006FastIA}, which may cause classical gradient-based approaches to fail or converge at unsatisfactory solutions, as depicted in Fig.~\ref{fig:M3500B}-(c).

 This work, shows that our proposed reinforcement learning (RL) agent can perform exceptionally well on graphs with poor initial guesses. The agent can be utilized to search and explore the pose trajectory space configurations by applying the optimal retractions~\cite{MANIF} directly to the desired poses, with the goal of returning and recovering the optimal global trajectory. In certain situations, obtaining ground truth trajectory labels in the SLAM domain may be laborious and expensive. The main advantage of leveraging a reinforcement learning-based agent is that existing supervised networks require to be trained on a large collection of labeled data. Suitable or empirical datasets used for label-based training are not always guaranteed~\cite{Overview}.

Our contributions can be summarized as follows:
\begin{itemize}

\item The first DRL model to learn a policy that predicts the optimal orientation retractions~\cite{MANIF} from pose-graph observations for the application of planar PGO. The proposed modular encoder-agent architecture is comprised of two main components. The first component is a neural network initially proposed in~\cite{2D-Pose-Graph-Neural-Network-Classifier} for graph optimality classification. The second is a recurrent-based soft actor-critic (SAC)~\cite{rlalgorithms1} agent whose policy predicts the optimal orientation retractions~\cite{MANIF}.

\item A planar pose-graph environmental framework based on~\cite{minisam} and GUI for visualization during evaluation. The environment and the proposed agent will be made publicly available\footnote{\href{https://sites.google.com/view/rl-pgo}{https://sites.google.com/view/rl-pgo}} to encourage more research into developing and comparing other RL approaches in the pose-graph domain.

\item Extensive experiments on simulated and real-world benchmarks illustrate the accuracy of the proposed agent, and further demonstrate the generalization capability of the agent to graphs never seen before during training.
  
\end{itemize}

\par

The paper is organized as follows: Sec.~\ref{related} provides an overview of the related works. Sec.~\ref{Problem_Statement} includes preliminary definitions and related terminology. In Sec.~\ref{Proposed_Approach}, we present our proposed environment, and describe the encoder and agent architecture. Sec.~\ref{Experimental_Results} and~\ref{Conclusion_and_future_works} present the experimental results, followed by the conclusions and future works.

\section{Related Work}
\label{related}

\par 

\textbf{Conventional Pose-Graph Optimization Approaches}. Popular MAP iterative solvers such as $\mathrm{g}^{2} \mathrm{o}$~\cite{g2o}, Ceres~\cite{Ceres}, and GTSAM~\cite{GTSAM} can be explicitly utilized to tackle problems that can be represented as a graph using nonlinear optimization methods. Line search methods such as Stochastic Gradient Descent and Gauss-Newton (GN) are typically used to perform nonlinear optimization. Given an initial guess, second order methods repeatedly  linearize a nonlinear least-squares objective function and then solve the unique least-squares equations until convergence~\cite{FactorGraphBook}. Depending on the quality of this initial guess, there is no assurance for convergence to global optimal solutions~\cite{LAGO}. As the non-convexities of the problem increase, iterative methods such as Powell's dogleg, Levenberg-Marquardt (LM), and Gauss-Newton, are subject higher computational efforts required due to recurring matrix computations involved in the linearization step. This in turn, led to an increased interest in regards to direct and indirect linear solvers as well as factorization techniques to be introduced. Direct and indirect sparse linear solver methods include the preconditioned conjugate gradient (PCG), SuiteSparseQR~\cite{SUITEQR}, and CHOLMOD~\cite{CHOLMOD}. For larger scaled problems, bundle adjustment methods were introduced to perform at a much more accelerated rate. This is done by performing computations synchronously parallel, distributing the workload, and making use of the recent developments in hardware, i.e, IPUs. An example of this was implemented in \cite{FutureMapping2} and \cite{Graphcore}.

\par
Bootstrapping or initialization strategies were also shown to improve convergence of iterative methods. Multi Ancestor Spatial Approximation Tree for 2D and 3D pose-graph optimization \cite{MASAT} demonstrate a computationally efficient and light-weight initialization method, which when followed by a gradient-based solver, achieves good results on classical 2D and 3D pose-graph optimization datasets. 

\par 
In recent years, the semidefinite relaxation (SDR) technique has been involved in exciting developments in the area of SLAM, and has also shown great significance on a variety of applications. In general, it can be applied to many non-convex quadratically constrained quadratic programs (QCQPs) such as the pose SLAM and Landmark SLAM problem instances. 


SE-Sync~\cite{SE-Sync} is a certifiably correct algorithm for performing synchronization over the special Euclidean group. Leveraging duality to devise an algorithm was also shown to enable global verification of a given estimate~\cite{2d_Duality}. The sparse bounded degree sum-of-squares (Sparse-BSOS) optimization method~\cite{Sparse-BSOS}, formulate SLAM problems as polynomial optimization programs and demonstrate the ability to achieve global minimum solutions without initialization. A deep learning approach for pose-graph global optimality classification was also recently proposed in~\cite{2D-Pose-Graph-Neural-Network-Classifier}.

\par
\textbf{Learning-Based Optimization Approaches}. There has been a recent surge of interest on incorporating deep neural networks into state estimation and SLAM pipelines.
Others have approached alternative solutions which are learned variations of the traditional approaches. LS-Net~\cite{clark2018lsnet} introduced the first approach to a learned optimizer for minimizing photometric residuals. Several learning-based methods involve reparametrization to allow for gradient back propagation and end-to-end learning. An example of this was illustrated in~\cite{Ba-Net} and~\cite{Grad-SLAM}, which involved a reparametrization of the damping mechanism to enable differentiability in Levenberg-Marquardt solvers. Recent line of works, present an end-to-end approach for learning estimators modeled as factor-graph smoothers for state estimation applications. A surrogate loss was proposed in~\cite{Learned_smoothers} for end-to-end training of smoothing-based estimators, and was evaluated on Visual Tracking, and Visual Odometry tasks.

\par
\textbf{Graph Neural Networks.} There has also been a plethora of works involving Graph Neural Networks for multiple rotation averaging (MRA). NeuRoRA~\cite{NeuRoRA} were the first to apply a graph neural network to regress the rotations in a view graph. The model consisted of a cleaning network which was responsible for mitigating outlier measurements, followed by an additional network used for fine tuning and regressing the absolute rotations given the measurements. Using a single Message Passing Neural Network (MPNN) with edge attention, \cite{RotationAttention} further improved the robustness, training time, and execution speed of the previous deep learning model and eliminated the requirement for an additional network. Recently, PoGO-Net~\cite{Pogo} introduce a novel joint loss MRA formulation which was shown to outperform state-of-the-art and operate in real-time.

 
\par
\textbf{Reinforcement Learning.} 
All the aforementioned learning techniques either require ground truth labels or apply techniques to enable differentiability for end-to-end training. 
Reinforcement learning revolves around the concept of learning from interaction~\cite{RLTextbook}. It is different from the supervised and unsupervised machine learning paradigms in that it does not have an external supervisor to pair situations with labeled actions and does not try to propose a valid structure hidden in unlabelled situations. Instead, RL mainly focuses on maximizing a reward signal through interactions between the agent and an environment. It is essentially based on associating transitional experiences to actions in such a manner that would maximize a reward signal, and is used in several applications, e.g. healthcare~\cite{HealthcareRL}, robotics~\cite{RoboticsRL}, finance~\cite{FinanceRL}, and many more~\cite{DeepRLBOOK}.

\section{Problem Statement}
\label{Problem_Statement}
In Sec.~\ref{Lie_theory}, we provide relevant background knowledge and preliminary definitions in regards to PGO and Lie Theory. Related RL terminology is also further provided in Sec.~\ref{RL_terminology}. 

\subsection{2D Pose-Graph Optimization and Lie Theory}\label{Lie_theory}
Assuming independent Gaussian measurement noise for relative orientation and translation measurements, i.e.  $\boldsymbol{\Sigma}_{i j}=\operatorname{diag}\left(\boldsymbol{\sigma}_{t}, \boldsymbol{\sigma}_{R}\right)$, we formulate the problem for planar scenarios as follows:
Given $m$ relative pose measurements encoded by edges $(i, j) \in \mathcal{E}$ on a directed graph shared between $n$ robot poses, and co-variance matrix for all measurement pairs, the goal of pose-graph optimization is to return the optimal estimate or configuration of poses which best fit the measurements observed. This can be more formally denoted as seeking the minimum of the objective cost function $F(\boldsymbol{x})$~\cite{LAGO}:


\begin{equation}\label{eq:1}
\begin{aligned}
F(\boldsymbol{x}) &=\sum_{(i, j) \in \mathcal{E}}\left\|\boldsymbol{R}_{i}^{\top}\left(\boldsymbol{t}_{j}-\boldsymbol{t}_{i}\right)-\widetilde{\boldsymbol{t}}_{i j}\right\|_{{{}\boldsymbol{\Sigma}_{\sigma}}_{t}}^{2}+\\
&\sum_{(i, j) \in \mathcal{E}}\left\|\left(\boldsymbol{R}_{j}-\boldsymbol{R}_{i}\right)-\widetilde{\boldsymbol{R}}_{i j}\right\|_{\boldsymbol{\Sigma}_{\sigma_{R}}}^{2},
\end{aligned}
\end{equation}

\noindent where $\|.\|_{\Sigma}^{2}$ symbolizes the  Mahalanobis distance, and $F(\boldsymbol{x})$ is a unitless weighted squared sum of all error residuals used to determine the accuracy of the state estimation (smaller the better). $\boldsymbol{x}\in \mathbb{S E}(2)$ denotes the set of all incremental pose coordinates. For each pose in the set, $\boldsymbol{x}_{i}=\left[\boldsymbol{t}_{i}^{\top} \boldsymbol{R}_{i}\right]^{\top}$, $\boldsymbol{t}_{i}\in \mathbb{R}^{2}$ and $\boldsymbol{R}_{i}\in \mathbb{S O}(2), \forall i=1 \ldots n$ denote the absolute translation and orientation configurations. The relative translation and orientation measurements observed between neighboring poses $i$ and $j$ are represented as $\widetilde{\boldsymbol{t}}_{i j}\in \mathbb{R}^{2}$ and $\widetilde{\boldsymbol{R}}_{i j}\in \mathbb{S O}(2)$. The level of difficulty in solving this optimization task lies in the estimation of the absolute orientations. It was shown in~\cite{Initialization} and~\cite{LAGO} that estimating rotations first in the case of pose-graph initialization, allows for greater convergence guarantees and closed form solutions in the planar case. Further proven in~\cite{LAGO}, the pose-graph optimization would be a linear least-squares problem if orientation states are known which allows for significant computational advantage. This principal is illustrated in our proposed method where a reinforcement learning agent seeks to recover the optimal orientation configurations for each robot pose in $\mathbb{S O}(2)$, followed by a linear least-squares translation estimation step shown in \cite{LAGO}. When applying perturbations on the orientation states, one cannot just simply apply matrix addition, as this operation does take into consideration the wrap-around in the $\mathbb{S O}(2)$ Lie Group. Instead, we apply updates via local reparametrization more commonly known as the exponential mapping, followed by the retraction operation which is denoted by the symbol $\oplus$~\cite{MANIF}. In $\mathbb{S O}(2)$, this is defined as

\begin{equation}\label{eq:2}
\boldsymbol{R}_{i} \oplus \xi \triangleq \boldsymbol{R}_{i} \cdot \exp \hat{\xi},
\end{equation}

\noindent where $\xi \in \mathbb{R}$, and the hat operator is responsible for the transformation from Cartesian vector space to the Lie algebra, which is a 2$\times$2 matrix. The inverse of the exponential mapping is referred to as the logarithmic mapping, which transfers elements from the Lie group to Cartesian coordinates denoted as: $\text{Log} \left(\boldsymbol{R}_{\boldsymbol{i}}\right)$.

\subsection{Reinforcement Learning Definitions}\label{RL_terminology}
The basic components of reinforcement learning consist of an agent and an environment. The main objective of a reinforcement learning algorithm involves training an agent to learn an optimal policy $\pi(a|s)$, such that the agent can achieve a high cumulative reward under a user defined evaluation metric, known as the reward function. A policy is defined as a probabilistic mapping from states to actions. For every discrete time step of interaction $t$, the agent is responsible for making optimal sequential decisions by applying actions $\mathbf{a}_{t} \in \mathcal{A}$, observing a state $\mathbf{s}_{t} \in \mathcal{S}$, and then receiving a feedback signal or reward $R\left(\mathbf{s}_{t}, \mathbf{a}_{t}\right)$ for the corresponding state-action pair. We also define a $Q$ value function which returns the expected sum of rewards assuming the agent is in state $\mathbf{s}$ and performs actions $\mathbf{a}$ following policy $\pi$.



The standard RL objective seeks to maximize the expected sum of rewards: $\mathbb{E}_{\pi}\left[\sum_{t} \gamma^{t} R\left(\mathbf{s}_{t}, \mathbf{a}_{t}\right)\right]$, where $\gamma \in[0,1)$ is the weighting which defines how much importance we give for future rewards. It is also assumed that the state must include information about all aspects
of the past interactions which enables informative decisions for the future~\cite{RLTextbook}. When an observation does not contain the complete state information, the environment is said to be partially observable.

\section{Proposed Approach}\label{Proposed_Approach}

\begin{figure*}
    \centering
    \includegraphics[width=0.9\textwidth]{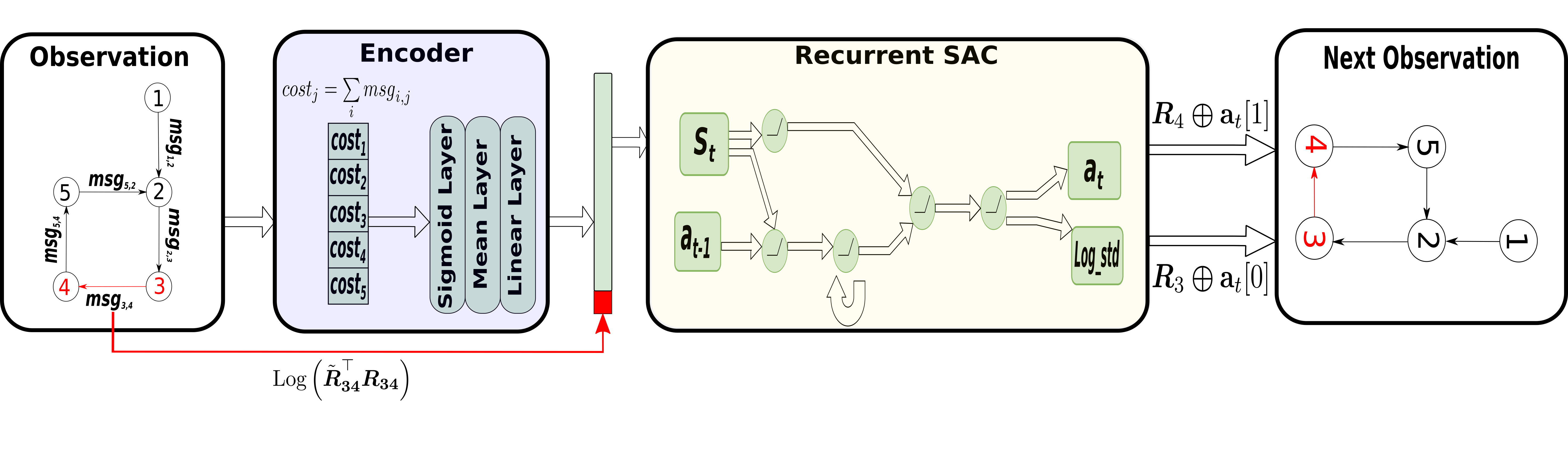}
    \vspace{-8 mm}
    \caption{Agent (red), cycling through every edge in the pose-graph and applying optimal retractions to the orientation components of each neighbouring pose. }
    \label{fig:agent_environment}
    \vspace{-5 mm}
\end{figure*}

The end-to-end agent-environment interaction is illustrated in Fig.~\ref{fig:agent_environment}. For each episode step, the orientation residual components of the input graph are first passed into the encoder network. The encoder then returns a highly expressive low dimensional input representation of the state. Once the state is then provided to the recurrent-based SAC policy network, the optimal retractions on neighbouring poses are applied. In Sec.~\ref{Environment_Details} and~\ref{Encoder_Architecture}, we review the environment details and the framework used to encode our observations into a descriptive latent embedding. Sec.~\ref{Recurrent_SAC} overviews the recurrent SAC policy network used to apply the optimal retractions on the graph orientation state space.

\subsection{Environment Details}\label{Environment_Details}
 To frame the pose-graph optimization problem in terms of reinforcement learning, we need to first define an environment. The environmental framework was inspired from the synthetically generated pose-graphs provided in~\cite{LAGO}. Each pose-graph instance is categorized by five environmental parameters: the number of poses $n$, orientation measurement uncertainty $\boldsymbol{\sigma}_{R}$, translation measurement uncertainty $\boldsymbol{\sigma}_{t}$, inter-nodal distance spacing $d$, and probability of loop closures $lc$.
 \par
 
 As shown in Fig.~\ref{fig:agent_environment}, the agent's location along the graph is highlighted in red, where steps occur between each edge. The number of steps per episode is dependant on the number of edges in the graph and a user defined value referred to as the number of cycles. This value determines the number of times the agent traverses across each edge, and termination of an episode. The definition of one cycle consists of when an agent traverses every edge in the pose-graph once.

 For each step-by-step transition along the edges, the output from the policy is passed through the $tanh$ activation layer and then multiplied by a user defined action range factor. The resultant vector is utilized to apply a retraction on the neighbouring poses connected by the edge, in which the agent resides at that particular instance in time: $R_{i} \oplus a_t[0]$ and $R_{j} \oplus a_t[1]$. In one cycle, each pose is perturbed twice assuming no loop closures. Following the completion of an episode, the final orientation estimate returned by the agent is then utilized in the concluding linear least-squares translation estimation provided in~\cite{LAGO}.
 
 The state in this case is comprised of two components. The first component includes the encoded history of the orientation residuals at each time step. The second component is essentially the angular difference between the observed and measured orientations corresponding to the edge in which the agent is located at time $t$:
\begin{equation}\label{eq:difference_for_state}
\text{Log} \left(\tilde{\boldsymbol{R}}_{\boldsymbol{i j}}^{\top} {\boldsymbol{R}}_{\boldsymbol{i j}}\right)\text{,}
\end{equation}

\noindent where ${\boldsymbol{R}}_{i j} \doteq \boldsymbol{R}_{i}^{\top} \boldsymbol{R}_{j}\text{,} \,\, (i, j) \in \mathcal{E}$. The graph observation encoding details are further discussed in Sec.~\ref{Encoder_Architecture}.
 
 The reward function for each time step is hybrid (dense/sparse) and is calculated as
\begin{equation}
Reward= \frac{100}{OC+1} \text{,}
\end{equation}
\noindent with an additive constant of $+25$ for every step the function output experiences a relative decrease by factor of $10$, i.e. (0.001, 0.0001,0.00001). The orientation cost $(OC)$ is
 
 \begin{equation}\label{eq:1}
\begin{aligned}
OC = \sqrt{\sum_{(i, j) \in \mathcal{E}}\left\|\widetilde{\boldsymbol{R}}_{i j}-\boldsymbol{R}_{i j}\right\|_{F}^{2}}\quad\text{,}
\end{aligned}
\end{equation}
where $\|\boldsymbol{B}\|_{F}$, denotes the Frobenius norm of matrix $\boldsymbol{B}$.

\subsection{Graph Encoder Architecture}\label{Encoder_Architecture}
We adopt the architecture proposed in~\cite{2D-Pose-Graph-Neural-Network-Classifier}, formally presented for the task of global optimality prediction. The poses or in other words, nodes of each graph input, store a cost feature and the absolute orientation $\boldsymbol{R}_{i}$ of the node itself. We utilize an augmented message passing function which is dependant on orientation components only, as opposed to both translation and orientation as depicted in Eq.~\eqref{eq:msg}. Further provided by~\cite{2D-Pose-Graph-Neural-Network-Classifier}, once the input pose-graph observation passes forward, a two step process occurs. The first step is the message passing step which involves computing the Frobenius norm of the absolute orientations shared by each edge, where $\beta$ is a learnable weight:

\begin{equation}\label{eq:msg}
m s g_{i, j}=\beta\times\left\|\boldsymbol{R}_{i} \boldsymbol{R}_{i j}-\boldsymbol{R}_{j}\right\|_{F}.
\end{equation}

Proceeding the computation of messages for each of the connected nodes, an aggregate sum is then stored in the cost feature associated with the corresponding node itself, 
\begin{equation}\label{eq:cost}
{cost}_{j}=\sum_{i}{m s g}_{i,j}.
\end{equation}

The mean of all cost features from graph observation at time $t$ is then passed through a linear layer and concatenated with the angular difference between the observed and measured orientations, corresponding to the agent's edge location computed from Eq.~\eqref{eq:difference_for_state}.
The output dimension of the linear layer is user-defined and for all of our evaluations we set this value to $20$. This resultant state vector $\mathbf{s}_{t}$ at time $t$ is passed into the recurrent SAC policy for an action to be applied.

\subsection{Recurrent SAC}\label{Recurrent_SAC}
\par

Soft actor-critic (SAC)~\cite{SAC_Original}, \cite{SAC1} is a state-of-the-art continuous control RL algorithm with an augmented objective:

\begin{equation}\label{eq:SAC_obj}
\underset{\pi}{\mathbb{E}}\left[\sum_{t} \gamma^{t}\left[R\left(\mathbf{s}_{t}, \mathbf{a}_{t}\right)+\alpha \mathcal{H}\left(\pi\left(\cdot \mid \mathbf{s}_{t}\right)\right)\right]\right]\text{,}
\end{equation}
\noindent where the temperature parameter $\alpha$ dictates the relative importance between the entropy $\mathcal{H}$ and reward $R$. Thus, the agent is encouraged to explore the state space and unseen trajectories which can speed up learning and prevent sub optimal convergence.
We utilize a recurrent version of the original implementation as shown in~\cite{rlalgorithms1}. After observing the current state and history of previous state-action pairs, the recurrent SAC policy network applies the optimal retractions to each of the poses shared by the edge in which the agent is located at for every time step. 

The recurrent SAC algorithm utilizes two networks, the Q-function and the policy. We will consider the policy $\pi_{\phi}$, parameterized by $\phi$ modelled as a Gaussian with mean $\mathbf{a}_{t}$, and covariance as depicted in Fig.~\ref{fig:agent_environment}. The two dimensional action vector at time $t$ is given by:
\begin{equation}\label{eq:19}
\mathbf{a}_{t}=\pi_{\phi}\left(\epsilon_{t} ; \mathbf{s}_{t},\mathbf{a}_{t-1},\mathbf{z}_{t}\right) \text{,}
\end{equation}

\noindent where $\epsilon_{t}$ is an input noise vector sampled from a normal distribution with mean 0, and standard deviation of 1. Both policy and Q value architectures follow the structure of the recurrent-based DDPG algorithm proposed in~\cite{Sim-to-real1}.

The recurrent branch further allows the agent to make informed optimal decisions on the next actions from previous state and action pairs, by the internal state representation~\cite{Sim-to-real1}: $z_{t}=z\left(h_{t}\right)$ modelled by an LSTM. We denote $h_{t}=\left[a_{t-1}, s_{t-1}, a_{t-2}, s_{t-2}, \ldots\right]$ to represent the history of the previous states and actions.

The objective loss functions are essentially the same as presented in~\cite{SAC1}, with slight modification to the gradient estimators, accounting for back propagation through time (BPTT). Details can be referred to in~\cite{Sim-to-real1} and~\cite{heess2015memorybased}.

\section{Experimental Results}\label{Experimental_Results}
In this section, we evaluate our RL approach and provide comparisons against the gradient-based $\mathrm{g}^{2} \mathrm{o}$~\cite{g2o} Gauss Newton, and Levenberg-Marquardt iterative solvers.
As proven in~\cite{LAGO}, the non-convexity of a given problem instance are related to the ratio of orientation to translation uncertainty $\frac{\boldsymbol{\sigma}_{R}}{\boldsymbol{\sigma}_{t}}$, and the squared sum of all measurement distances. 
In Sec.~\ref{Ratio_test} and~\ref{Inter_test}, we conduct analytical tests similarly done and inspired by~\cite{LAGO}, where comparisons were made against Gauss Newton $10$ (GN10) and $5$ (GN5) iterations in their evaluations. For our comparisons we set the maximum to $100$. The quality of estimation provided by our approach is assessed under the influence of the adjustable environmental parameters, and further illustrate the effectiveness of the proposed approach under challenging scenarios. A head-to-head comparison with the Levenberg-Marquardt solver is also provided on standard real-world and synthetic benchmarks in Sec.~\ref{Standard_benchmark_test}. 
 
 \par
\begin{figure}[ht!]
\centering

\subfloat[\scriptsize Cumulative reward for training environments 1-3]{\includegraphics[width=0.45\textwidth]{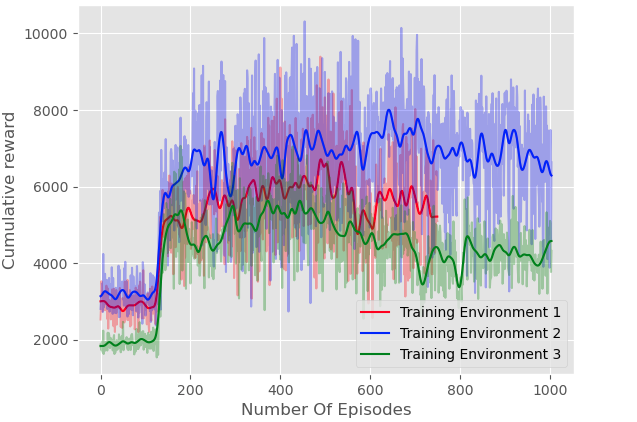}} 
\hfill
\subfloat[\scriptsize Cumulative reward for training environments 4-5]{\includegraphics[width=0.45\textwidth]{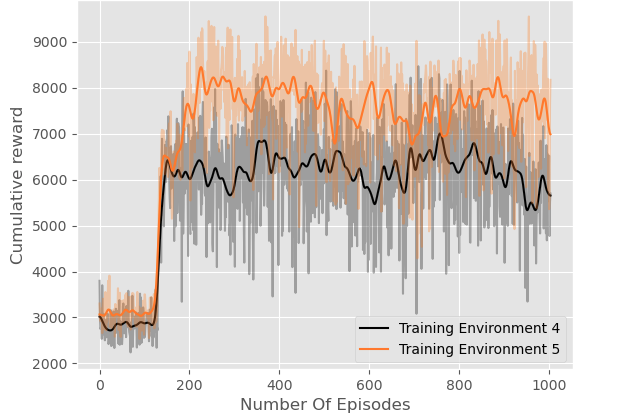}} \\

\caption{\scriptsize (a)-(b) Cumulative reward plots for Training Environments 1-5. \label{fig:reward_plots}}
\vspace{-7 mm}
\end{figure}

 We train five separate agents, and for all evaluations depict results for the agent which performed best on the testing environment. Each of the agents are trained on small pose-graph environments of size $n=20$ randomly sampled from their assigned environmental noise distributions every episode. The agents are then evaluated on synthetic and real-world graphs, much larger in size and unseen during training to demonstrate generalizability. The five training environment details are indicated in Table~\ref{table:Training_environment_list}, and cumulative reward plots for these agents during training are depicted in Fig.~\ref{fig:reward_plots}.
 
\begin{table}[h]
\centering
\begin{tabular}{|c|c|c|c|c|c|c|c|}
\hline
\multirow{2}{*}{Env.} & \multirow{2}{*}{$n$} & \multirow{2}{*}{$\boldsymbol{\sigma}_{R}$~[rad]} & \multirow{2}{*}{$\boldsymbol{\sigma}_{t}$~[m]} & \multirow{2}{*}{$d$~[m]} & \multirow{2}{*}{$lc$} & \multirow{2}{*}{Cycles} & Action\\
& & & & & & & Range \\
\Xhline{3\arrayrulewidth}
$1$ & $20$ & $0.3$ &  $0.01$ & $3$ & $0.5$ & $7$ & $0.25$\\
\hline
$2$ & $20$ & $0.3$ &  $0.01$ & $3$ & $0.5$ & $8$ & $0.25$\\
\hline
$3$ & $20$ & $0.2$ &  $0.1495$ & $1$ & $0.5$ & $5$ & $0.4$\\
\hline
$4$ & $20$ & $0.2$ &  $0.1495$ & $1$ & $0.5$ & $6$ & $0.25$\\
\hline
$5$ & $20$ & $0.1$ &  $0.01$ & $10$ & $0.5$ & $6$ & $0.25$\\
\hline
\end{tabular}
\caption{Training environment parameters include the number of poses $n$, orientation measurement uncertainty $\boldsymbol{\sigma}_{R}$, translation measurement uncertainty $\boldsymbol{\sigma}_{t}$, inter-nodal distance spacing $d$, and probability of loop closures $lc$. See Sec.~\ref{Environment_Details} for more details.}
\label{table:Training_environment_list}
\vspace{-7 mm}
\end{table}

 \subsection{Implementation and Training Details}
 For the policy and Q-value network, all related layers consist of $512$ fully connected units followed by $512$ LSTM units in the recurrent branch. Kaiming Initialization~\cite{Kaiming} was employed on both Q-function and policy network weights. Rectified non-linearity~\cite{ReLU} (ReLU) were used for all hidden layers. We utilized Adam optimizer~\cite{kingma2017adam} with a learning rate of ${3.0E-04}$, minibatch size of $128$ and a discount factor of $\gamma=1.0$. All Q-function and policy networks are updated after every episode with target networks updated by an exponential moving average, with $\tau=1.0E-02$. The networks are trained on a single Nvidia GeForce RTX 3090 GPU with 24GB memory. For best performance the final policy was set to deterministic and all objective cost function values $F(x)$, as well as optimization times recorded, are averaged over $10$ runs or episodes.

 \subsection{Effect on Measurement Uncertainty Ratio}\label{Ratio_test}
\begin{figure}[t!]
 \centering
\subfloat[Ground Truth trajectory (green) and initial noisy estimate (red)]{\includegraphics[width=0.14\textwidth]{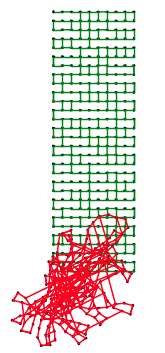}} 
\hfill
\subfloat[GN100 estimate]{\includegraphics[width=0.17\textwidth]{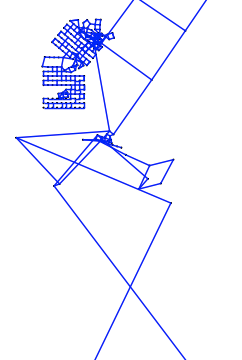}}
\hfill
\subfloat[RL (black) and  RL+GN50 (magenta) estimates]{\includegraphics[width=0.17\textwidth]{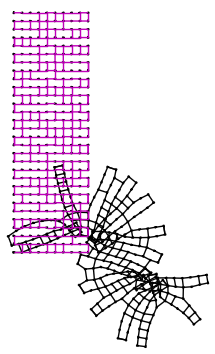}} \\

\caption{Analysis on the influence of measurement uncertainty ratio for the instance $\boldsymbol{\sigma}_{t}=0.05m$. The RL+GN50 estimate (magenta) as shown in (c), only required 8 iterations for convergence and successfully recovered the global minimum solution. As the ratio of uncertainty increases, the RL standalone estimate (black) provides a much accurate estimation when compared to GN100 (blue).}
\label{fig:ratio_test_fig}
\vspace{-5 mm}
\end{figure}

 In this analytical case study, we allow Gauss Newton to perform $100$ iterations (GN100) which can be interpreted as applying 100 pose retractions or {\it actions} to each pose in the graph. We utilize the agent trained in environment number $2$ where the number of evaluation cycles was set to $8$, and therefore performs far fewer retractions per pose than GN100.
 This analysis involves evaluation of the performances on a test pose-graph environment with decreasing translation uncertainty while all other parameters are kept fixed. In this case, $n=300$, $\boldsymbol{\sigma}_{R}=0.3$rad, $d=3$m, $lc=0.5$, while $\boldsymbol{\sigma}_{t}$ ranges from $[0.2, 0.1, 0.05, 0.03, 0.01]$m. Objective cost function values and total elapsed optimization times are presented in Table~\ref{table:uncertainty_ratio} and an example of the $\boldsymbol{\sigma}_{t}=0.05$m instance is shown in Fig.~\ref{fig:ratio_test_fig}. It is observed that as the ratio of orientation to translation uncertainty increases, estimates produced by GN100 are extremely inaccurate as the non-convexities of the pose-graph optimization problem also increase. Standalone RL was shown to outperform GN100 in situations where $\boldsymbol{\sigma}_{t}=0.05$m, $0.03$m, and $0.01$m. It is also noticed when utilizing the RL estimate as an initial guess, the ground truth trajectory and hence, global minimum, is returned in almost all cases in far fewer retractions per pose. Regardless of GN being set to perform $50$ iterations, once bootstrapped by the RL estimate graph instances $\boldsymbol{\sigma}_{t}=0.2$m, $0.1$m, $0.05$m, and $0.03$m converged in less than $11$.
 
\begin{table}[!h]
\centering 
\begin{tabular}{|c|l?c|c|c|}
\hline
$\boldsymbol{\sigma}_{t}$~[m] & Metric & GN100 & RL & RL+GN50 \\

\Xhline{3\arrayrulewidth}
\multirow{2}{*}{$0.2$} 
     & $F(x)$     & \textbf{0.00} & 6.04E+03   & \textbf{0.00} \\\cdashline{2-5}
     & time [s] & 0.04          & 10.08      & 10.12 \\\hline
\multirow{2}{*}{$0.1$} 
     & $F(x)$  & \textbf{0.00}  & 2.68E+04    & \textbf{0.00} \\\cdashline{2-5}
     & time [s] & 0.04         & 9.87       & 9.91           \\\hline
\multirow{2}{*}{$0.05$} 
     & $F(x)$     & 1.59E+08      & 5.31E+04  & \textbf{0.00} \\\cdashline{2-5}
     & time [s] & 0.04         & 9.69      & 9.73 \\\hline
\multirow{2}{*}{$0.03$} 
     & $F(x)$     & 5.76E+06      & 1.89E+05  & \textbf{0.00} \\\cdashline{2-5}
     & time [s] & 0.04         & 10.01     & 10.05 \\\hline
\multirow{2}{*}{$0.01$} 
     & $F(x)$     & 1.47E+10      & 5.52E+06  & \textbf{1.21E+04} \\\cdashline{2-5}
     & time [s] & 0.04         & 9.97      & 10.01 \\\hline     
\end{tabular} 
\caption{Effect on measurement uncertainty ratio while other environmental parameters are held fixed.}
\label{table:uncertainty_ratio}
\vspace{-5 mm}
\end{table}

\subsection{Effect on Inter-nodal distance}\label{Inter_test}
 In this section we conduct our evaluations on test pose-graph environments synthetically generated with various inter-nodal distance spacing $d$, while all other parameters are kept fixed. The test environment parameters in this case are, $n=300$, $\boldsymbol{\sigma}_{R}=0.1$rad, $\boldsymbol{\sigma}_{t}=0.01$m, $lc=0.5$, while $d$ ranges from $[1, 3, 5, 8, 10]$m. Agent trained on environment $5$ was utilized for evaluations and the final metrics are provided in Table~\ref{table:inter-nodal_test}.

\begin{table}[t]
\vspace{1 mm} 
\centering 
\begin{tabular}{|c|l?c|c|c|}
\hline
$d$~[m] & Metric & GN100 & RL & RL+GN10 \\

\Xhline{3\arrayrulewidth}
\multirow{2}{*}{$1$} 
     & $F(x)$     & \textbf{0.00} & 6.82E+04   & \textbf{0.00} \\\cdashline{2-5}
     & time [s] & 0.04          & 6.96      & 7.0 \\\hline
\multirow{2}{*}{$3$} 
     & $F(x)$  & \textbf{0.00}  & 7.80E+05    & \textbf{0.00} \\\cdashline{2-5}
     & time [s] & 0.04         & 6.97       & 7.01           \\\hline
\multirow{2}{*}{$5$} 
     & $F(x)$     & 1.17E+06      & 1.03E+06  & \textbf{0.00} \\\cdashline{2-5}
     & time [s] & 0.04         & 6.96      & 7.0 \\\hline
\multirow{2}{*}{$8$} 
     & $F(x)$     & 1.58E+07      & 5.82E+06  & \textbf{0.00} \\\cdashline{2-5}
     & time [s] & 0.04         & 6.96     & 7.0 \\\hline
\multirow{2}{*}{$10$} 
     & $F(x)$     & 3.35E+14      & 1.24E+07  & \textbf{4.31E+05} \\\cdashline{2-5}
     & time [s] & 0.04         & 6.96      & 7.0 \\\hline     
\end{tabular} 
\caption{Effect on inter-nodal distance spacing while other environmental parameters are held fixed.}
\label{table:inter-nodal_test}
\vspace{-5 mm}
\end{table}

\begin{figure}[ht!]
 \centering
\subfloat[Ground Truth trajectory (green) and initial noisy estimate (red)]{\includegraphics[width=0.18\textwidth]{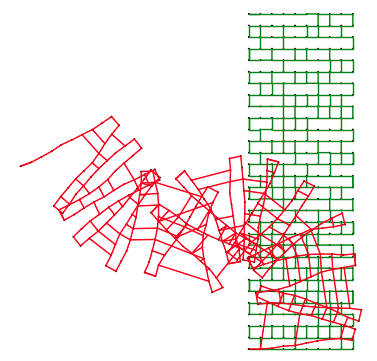}} 
\hfill
\subfloat[GN100 estimate]{\includegraphics[width=0.25\textwidth]{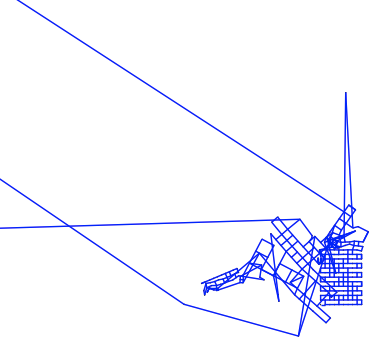}}
\hfill
\subfloat[RL (black) and  RL+GN10 (magenta) estimates]{\includegraphics[width=0.18\textwidth]{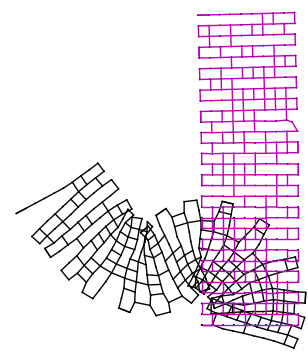}} \\

\caption{Analysis on the influence of inter-nodal distance spacing for the instance $d=10$m. The RL+GN10 estimate (magenta) as shown in (c), only required 8 iterations for convergence, to a visually meaningful solution. As the sum of of total measurement distances increase, the proposed RL agent (black) provides a much accurate estimation when compared to GN100 (blue).}
\label{fig:inter_test_fig1}
\vspace{-7 mm}
\end{figure}

One may notice that standalone RL outperforms GN100 on instances with larger inter-nodal distance spacing. These results also conform to the analysis performed in~\cite{LAGO} where it was further indicated that the inter-nodal distances are directly related to the sum of square measurements distances, which further increase the non-convexities of the pose-graph problem. GN10 bootstrapped by the RL estimate was also capable of achieving the global minimum cost in almost all instances. An example of the most challenging case $d=10$m is illustrated in Fig.~\ref{fig:inter_test_fig1}.

\subsection{Standard Benchmark Datasets}\label{Standard_benchmark_test}

To further assess the efficacy of the proposed approach, we evaluate performance on standard real-world and synthetic benchmarks provided by~\cite{carlone2012angular} and~\cite{LAGO}, never seen by our agent. Comparisons are made with $\mathrm{g}^{2} \mathrm{o}$'s LM solver fixed at $30$ (LM30), and $100$ (LM100) iterations, with stopping criteria based on reaching a relative error decrease or number of iterations. Additionally, we provide the objective cost function values and optimization time required for the LM30 estimate initially bootstrapped by our approach. The datasets include Manhattan world M3500 along with its three variants A, B, and C, which by default have standard deviations of $0.1$rad, $0.2$rad, and $0.3$rad added to the relative orientation measurements of the original. Further, we evaluate on City10K~\cite{iSAM}, as well as Intel and MIT, which were obtained by processing raw measurements from wheel odometry and laser range finder measurements obtained at the Intel Research Lab in Seattle and Killian Court. In this assessment, we double the number of cycles the agent is set to perform during test time except for the evaluation on City10K. Increasing the number of cycles for evaluations was found to further improve the quality of the RL estimate due to a larger number of pose refinements provided by the agent. Allowing the agent to perform more cycles however, come at the cost of longer optimization times required. This is apparent for graphs much larger in size such as City10K. In regards to evaluations on M3500A, B, C, City10k, and MIT, we employ the agent trained on environment number $3$ which was found to perform best on those datasets. For the M3500 and Intel datasets, agents trained on environments $4$ and $1$ were utilized for evaluation. Objective cost function values, and total elapsed optimization times are given in Table~\ref{table:real_world_comparisons}.

\begin{table}[t]
\vspace{1 mm} 
\centering 
\setlength\tabcolsep{2pt}
\begin{tabular}{|c|c?c|c|c|c|c|}
\hline
Dataset & \multirow{2}{*}{Metric} & \multirow{2}{*}{RL} & \multirow{2}{*}{LM30} & \multirow{2}{*}{LM100} & \multirow{2}{*}{RL+LM30} \\
{\scriptsize (\#\,of\,poses,\,\#\,of\,edges)}  & &  &  &  &  \\
\Xhline{3\arrayrulewidth}
M3500     & $F(x)$     & 1.97E+03 & \textbf{1.38E+02}   & \textbf{1.38E+02}   & \textbf{1.38E+02}      \\\cdashline{2-6}
{\scriptsize (3500, 5453)}     & time [s] & 870.02          & 0.26      & 0.46    & 870.28       \\\hline
M3500A     & $F(x)$     & 2.94E+04    & 1.46E+05   & 9.68E+04   & \textbf{5.72E+03}      \\\cdashline{2-6}
{\scriptsize (3500, 5453)}     & time [s] & 745.29          & 0.38      & 1.51    & 745.67       \\\hline
M3500B     & $F(x)$     & 4.66E+04    & 1.52E+04   & 1.43E+04   & \textbf{9.84E+03}      \\\cdashline{2-6}
{\scriptsize (3500, 5453)}     & time [s] & 744.70          & 0.33      & 1.48    & 745.03       \\\hline
M3500C     & $F(x)$     & 5.64E+04    & 6.01E+04   & 3.74E+04   & \textbf{7.12E+03}      \\\cdashline{2-6}
{\scriptsize (3500, 5453)}     & time [s] & 744.01          & 0.41      & 1.63    & 744.42       \\\hline
City10K     & $F(x)$     & 4.50E+04    & 4.53E+04   & 3.19E+04   & \textbf{5.12E+02}      \\\cdashline{2-6}
{\scriptsize (10000, 20687)}     & time [s] & 3914.25          & 2.72      & 9.85    & 3916.97      \\\hline
Intel     & $F(x)$     & 9.96E+05    & 1.30E+05   & 5.24E+04   & \textbf{4.65E+02}      \\\cdashline{2-6}
{\scriptsize (1228, 1483)}     & time [s] & 122.07          & 0.09      & 0.27    & 122.16       \\\hline
MIT     & $F(x)$     & 5.91E+03    & 1.11E+04   & \textbf{5.26E+02}   & 7.71E+02      \\\cdashline{2-6}
{\scriptsize (808, 827)}     & time [s] & 34.49          & 0.05      & 0.13    & 34.54       \\\hline
\end{tabular} 
\caption{Comparison Amongst Standard Benchmark Datasets.}
\label{table:real_world_comparisons}
\vspace{-5 mm}
\end{table}
\begin{figure}
\vspace{.5 mm}
\begin{minipage}[t]{0.01\columnwidth}
  \centering
  \end{minipage}
  \hfill
  \begin{minipage}[b]{0.3\columnwidth}
  \centering
        \caption*{RL}
  \end{minipage}
  \hfill
  \begin{minipage}[b]{0.3\columnwidth}
  \centering
    \caption*{LM100}
  \end{minipage}
  \begin{minipage}[b]{0.3\columnwidth}
  \centering
    \caption*{RL+LM30}
  \end{minipage}\\
\begin{minipage}[b]{0.01\columnwidth}
  \centering
    \caption*{\rotatebox{90}{ (a) M3500}}
  \end{minipage}
  \hfill
 \begin{minipage}[b]{0.3\columnwidth}
  \centering
    \includegraphics[width=1.05\linewidth]{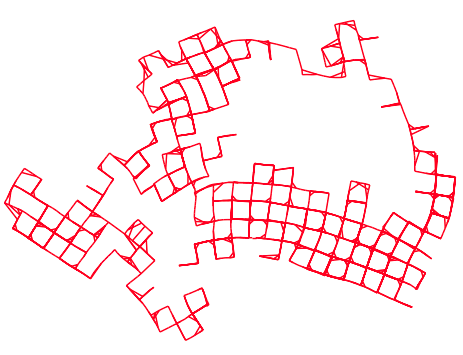}
  \end{minipage}
  \hfill
  \begin{minipage}[b]{0.3\columnwidth}
  \centering
    \includegraphics[width=1.05\linewidth]{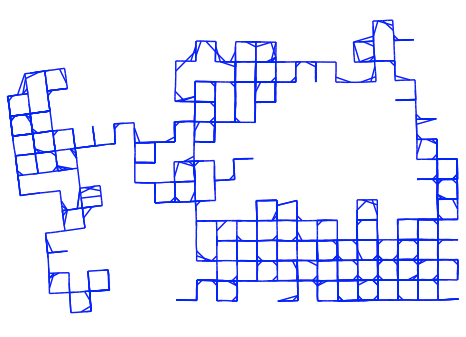}
  \end{minipage}
  \begin{minipage}[b]{0.3\columnwidth}
  \centering
    \includegraphics[width=1.04\linewidth]{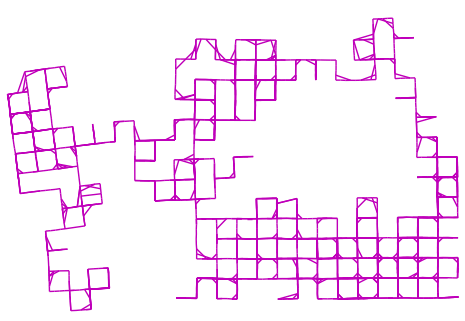}
  \end{minipage}\\
\begin{minipage}[b]{0.01\columnwidth}
  \centering
    \caption*{\rotatebox{90}{ (b) M3500A}}
  \end{minipage} 
   \begin{minipage}[b]{0.3\columnwidth}
  \centering
    \includegraphics[width=1.20\linewidth]{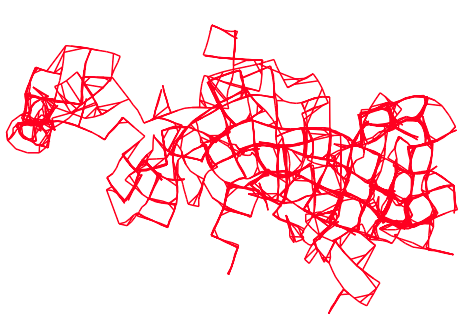}
  \end{minipage}
  \hfill
  \begin{minipage}[b]{0.3\columnwidth}
  \centering
    \includegraphics[width=1.20\linewidth]{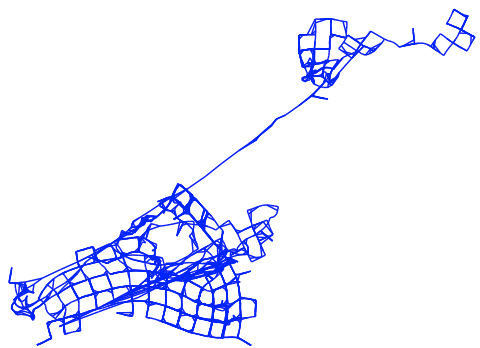}
  \end{minipage}
  \begin{minipage}[b]{0.3\columnwidth}
  \centering
    \includegraphics[width=1.11\linewidth]{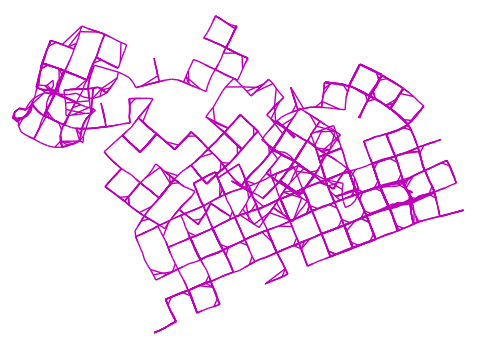}
  \end{minipage}\\
\begin{minipage}[b]{0.01\columnwidth}
  \centering
    \caption*{\rotatebox{90}{ (c) M3500C}}
  \end{minipage} 
   \begin{minipage}[b]{0.3\columnwidth}
  \centering
    \includegraphics[width=0.85\linewidth]{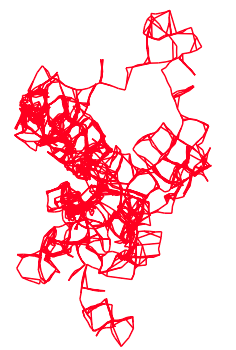}
  \end{minipage}
  \hfill
  \begin{minipage}[b]{0.3\columnwidth}
  \centering
    \includegraphics[width=0.85\linewidth]{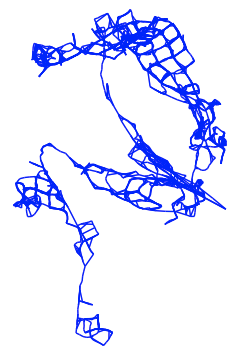}
  \end{minipage}
  \begin{minipage}[b]{0.3\columnwidth}
  \centering
    \includegraphics[width=0.85\linewidth]{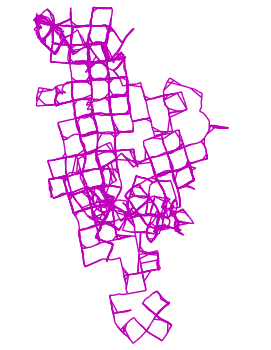}
  \end{minipage}\\
\begin{minipage}[b]{0.01\columnwidth}
  \centering
    \caption*{\rotatebox{90}{ (d) Intel}}
  \end{minipage}
   \begin{minipage}[b]{0.3\columnwidth}
  \centering
    \includegraphics[width=.9\linewidth]{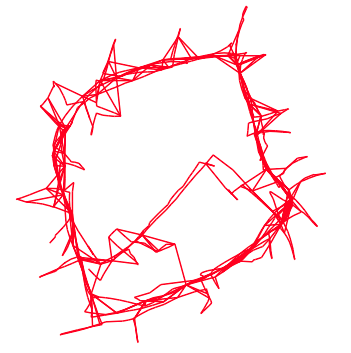}
  \end{minipage}
  \hfill
  \begin{minipage}[b]{0.3\columnwidth}
  \centering
    \includegraphics[width=1.1\linewidth]{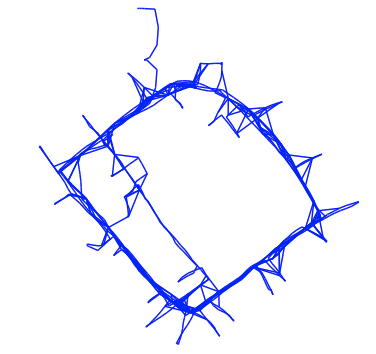}
  \end{minipage}
  \begin{minipage}[b]{0.3\columnwidth}
  \centering
    \includegraphics[width=.9\linewidth]{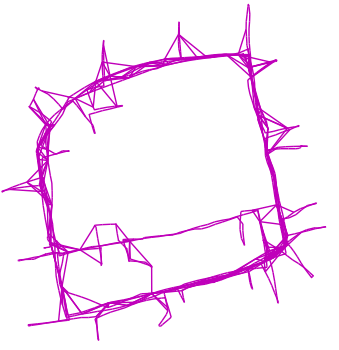}
  \end{minipage}\\
\begin{minipage}[b]{0.01\columnwidth}
  \centering
    \caption*{\rotatebox{90}{ (e) MIT}}
  \end{minipage}
   \begin{minipage}[b]{0.3\columnwidth}
  \centering
    \includegraphics[width=1.1\linewidth]{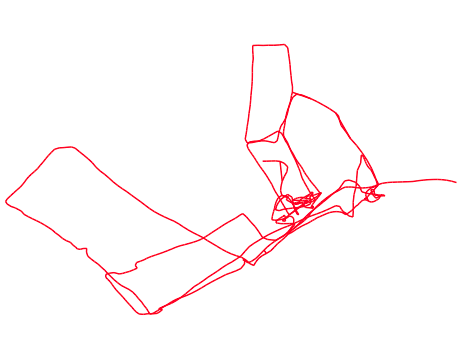}
  \end{minipage}
  \hfill
  \begin{minipage}[b]{0.3\columnwidth}
  \centering
    \includegraphics[width=1.1\linewidth]{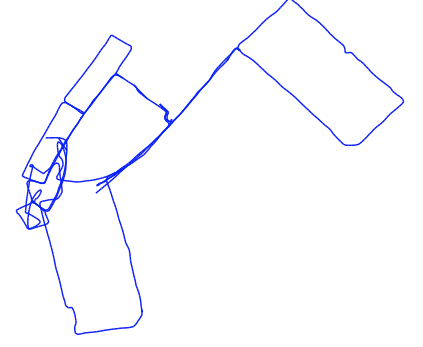}
  \end{minipage}
  \begin{minipage}[b]{0.3\columnwidth}
  \centering
    \includegraphics[width=1.0\linewidth]{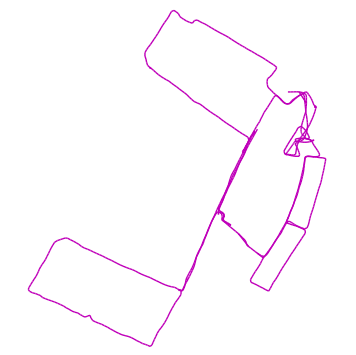}
  \end{minipage}\\
\begin{minipage}[b]{0.01\columnwidth}
  \centering
    \caption*{\rotatebox{90}{ (f) City10K}}
  \end{minipage}
   \begin{minipage}[b]{0.3\columnwidth}
  \centering
    \includegraphics[width=1.1\linewidth]{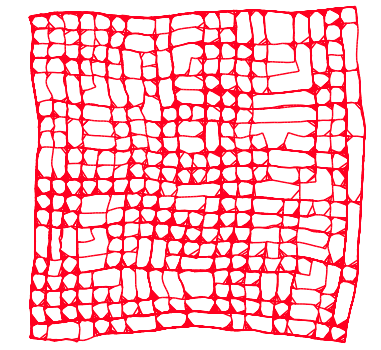}
  \end{minipage}
  \hfill
  \begin{minipage}[b]{0.3\columnwidth}
  \centering
    \includegraphics[width=1.1\linewidth]{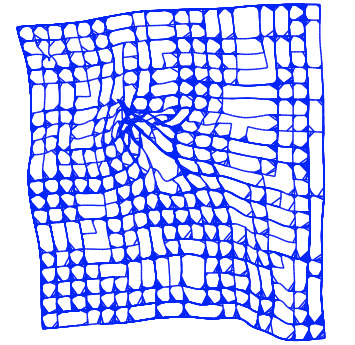}
  \end{minipage}
  \begin{minipage}[b]{0.3\columnwidth}
  \centering
    \includegraphics[width=1.05\linewidth]{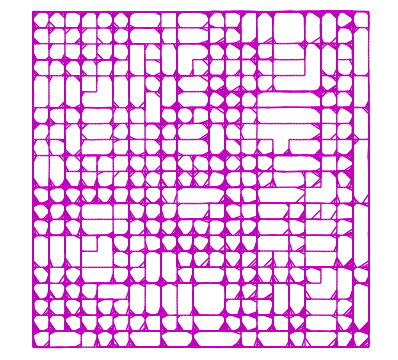}
  \end{minipage}
\caption{Comparison amongst the standard real-world and synthetic graphs. Performance metrics are depicted from Table~\ref{table:real_world_comparisons}. From left to right: Standalone RL best estimate from $10$ evaluations (red), LM100 estimate (blue), LM30 estimate when bootstrapped by the RL result as an initial guess (magenta). It is observed that RL standalone outperforms LM100 on M3500A. In all datasets except for MIT, RL+LM30 produced estimations with the highest quality (lowest objective cost function value). Interestingly, the standalone proposed RL agent was able to achieve solutions of adequate structure, despite having never seen any of these test graphs throughout training.  }
\label{fig:All_comparisons}  
\vspace{-5 mm}
\end{figure}  

It is observed that our proposed agent is effective in exploring the state space subject to highly non-convex cost functions. This is illustrated in instances M3500A, and C for example, which have a larger ratio of orientation to translation uncertainty due to the added standard deviation on the orientation measurements. In these graph instances the factors that influence non-convexity have a larger impact. In such cases, RL standalone was able to outperform LM30 in terms of state estimation accuracy. The RL estimate was also capable of achieving a smaller objective cost function value than LM100 on M3500A, further illustrated in Fig.~\ref{fig:All_comparisons}. Due to the sequential step-by-step nature of the agent applying only two actions at once, far much more computation time is required for the optimization episode to complete in all cases.

\par
In regards to dataset City10K, the number of edges in this graph and therefore, total sum of squared distance measurements, are also much larger than other datasets. Although RL standalone was able to outperform LM30 in City10K and MIT as well, by allowing more iterations, LM100 was eventually able to attain much accurate estimations. Nonetheless, when the RL estimate is utilized as a initial guess LM30 was capable of achieving even higher quality estimations than LM100 on all datasets except for MIT. We hypothesize that this is attributed to the fact that MIT has the fewest number of edges and loop closures, which provide the agent with less information on the residual state and decrease the number of actions applied. Further analysis which involved LM to run until convergence, indicate that the LM solver bootstrapped by our agent's estimate was still of higher quality (see website of the project: {\scriptsize \href{https://sites.google.com/view/rl-pgo}{https://sites.google.com/view/rl-pgo}}).

\section{CONCLUSION AND FUTURE WORKS}\label{Conclusion_and_future_works}
In this work, we have demonstrated a proposed RL agent to effectively explore the orientation state space in challenging pose-graph instances subject to highly non-convex cost functions for the application of planar pose-graph optimization. In particular scenarios our agent was capable of outperforming state-of-the-art Gauss Newton and Levenberg-Marquardt gradient-based solvers, in fewer retractions per pose. Nevertheless, although the RL-based approach was able to attain promising results, due to the small action space and inability to apply actions to all poses simultaneously the proposed agent does not seem to computationally scale well for graphs much larger in size. Thus, the methods presented in this work are more well-suited for offline Batch-SLAM applications. Our agent was shown to tolerate higher than usual levels of noise and inter-nodal distance spacing, which further degrades the quality of estimation. Comparisons on simulated environments further exploit the factors which influence non-convexity~\cite{LAGO}, where state-of-the-art gradient descent-based solvers may catastrophically fail or return poor quality estimations. As shown in Fig.~\ref{fig:reward_plots}, the pose-graph optimization problem can be modelled as a partially observable MDP.

In situations where obtaining ground truth trajectories or labels may be a laborious and expensive process, we have demonstrated and highlighted the agent's ability to perform well on much larger real-world and standard synthetic graphs, unseen during training. RL agents trained on smaller toy pose-graph environments of size $n=20$ were able to generalize well to graphs with dissimilar noise distributions and odometric trajectories at test time.

\par
Future works would involve extensions to pose-graph optimization instances in the 3D domain. Increasing action space size or a multi-agent framework may significantly reduce computational time. Moreover, a robust encoder architecture that is capable of mitigating the influence on outliers and false positive loop closures would allow for better standalone performance on real-world datasets. This may involve integrating robust kernels into the message passing function or de-noising layers adopted from~\cite{Pogo}.










\bibliographystyle{IEEEtran}
\bibliography{bib.bib}

\end{document}